\documentclass[twoside,twocolumn,10pt]{article}
\usepackage{wscg}           % includes a number of other packages (e.g., myalgorithm)
\RequirePackage{ifpdf}
\ifpdf
 \RequirePackage[pdftex]{graphicx}
 \RequirePackage[pdftex]{color}
\else
 \RequirePackage[dvips,draft]{graphicx}
 \RequirePackage[dvips]{color}
\fi
\usepackage[german,english]{babel}     % default = english
\usepackage{array}          % better tabular's & arrays, plus math tabular's
\usepackage{tabularx}      % for selfadjusting p-columns
\usepackage{nopageno}       % no page numbers at all; uncomment for final version
\usepackage{amssymb}
\usepackage{amsmath}
\usepackage{multicol}
\usepackage{multirow}
\usepackage{xspace}
\usepackage{hyperref}
\usepackage{tikz}

\usepackage{url}            %linebreaks in URL in References

\title{AI-based Density Recognition}
\author{
\parbox{0.45\textwidth}{\centering
Simone M{\"u}ller \\[1mm]
Leibniz Supercomputing Centre (LRZ)\\[1mm]
simone.mueller@lrz.de}
\hspace{0.3cm}
\parbox{0.45\textwidth}{\centering
Daniel Kolb\\[1mm]
Leibniz Supercomputing Centre (LRZ)\\[1mm]
daniel.kolb@lrz.de} \\
\\
\hspace{0.3cm}
\parbox{0.45\textwidth}{\centering
Matthias M{\"u}ller \\[1mm]
German Aerospace Center (DLR)\\[1mm]
matthias.mueller@dlr.de}
\hspace{0.3cm}
\parbox{0.45\textwidth}{\centering
Dieter Kranzlm{\"u}ller\\[1mm]
Ludwig-Maximilians-Universit{\"a}t (LMU)\\[1mm]
kranzlmueller@ifi.lmu.de}
}

\usepackage{url}
\makeatletter
\def\Uslash{\mathbin{\mathchar`\/}\@ifnextchar{/}{\kern-.15em}{}}
\g@addto@macro\UrlSpecials{\do \/ {\Uslash}}
\def\Ucolon{\mathbin{\mathchar`:}\@ifnextchar{/}{\kern-.1em}{}}
\g@addto@macro\UrlSpecials{\do : {\Ucolon}}
\makeatother
\begin{document}
\twocolumn[{
\csname @twocolumnfalse\endcsname
\maketitle  % full width title
%%%%%%%%%%%%%%%%%%%%%%%%%%%%%%%%%%%%%%%%%%%%%%%%%%%%%%%%%%%%%%%%%%%%%%%%%%%%%%%%%%%%%%%%%%%%%%%%%%%%
\begin{abstract}
\noindent
Learning-based analysis of images is commonly used in the fields of mobility and robotics for safe environmental motion and interaction. This requires not only object recognition but also the assignment of certain properties to them. With the help of this information, causally related actions can be adapted to different circumstances. Such logical interactions can be optimized by recognizing object-assigned properties. Density as a physical property offers the possibility to recognize how heavy an object is, which material it is made of, which forces are at work, and consequently which influence it has on its environment. Our approach introduces an AI-based concept for assigning physical properties to objects through the use of associated images. Based on synthesized data, we derive specific patterns from 2D images using a neural network to extract further information such as volume, material, or density. Accordingly, we discuss the possibilities of property-based feature extraction to improve causally related logics. 
\end{abstract}

%%%%%%%%%%%%%%%%%%%%%%%%%%%%%%%%%%%%%%%%%%%%%%%%%%%%%%%%%%%%%%%%%%%%%%%%%%%%%%%%%%%%%%%%%%%%%%%%%%%
\subsection*{Keywords}
AI, Density Recognition, Computer Vision
%%%%%%%%%%%%%%%%%%%%%%%%%%%%%%%%%%%%%%%%%%%%%%%%%%%%%%%%%%%%%%%%%%%%%%%%%%%%%%%%%%%%%%%%%%%%%%%%%%%
\vspace*{1.0\baselineskip}
}]
%%%%%%%%%%%%%%%%%%%%%%%%%%%%%%%%%%%%%%%%%%%%%%%%%%%%%%%%%%%%%%%%%%%%%%%%%%%%%%%%%%%%%%%%%%%%%%%%%%
\section{Introduction}
\copyrightspace

Modern machines and robots use various sensors to capture and navigate their surroundings. Particularly in road traffic, situations may appear inconspicuous at first sight but require constant attention and quick reactions. This can involve evaluating the potential risks in autonomous driving scenarios when a car not only recognizes objects but can also estimate the potential damage in the event of a collision and adapt its driving behavior accordingly. Additional information could increase a system's scope of action and decision-making as well as the automatic assessment of real-life scenes.

Whether it is a ball that rolls onto the road, a car that suddenly brakes, or an item that falls off a moving vehicle. Such reactions are often based on causal relationships that are logical for us humans but not for machines. Despite their logic, machines lack the necessary background knowledge and specific skills, such as the assessment of physical properties, to gain a causal understanding.

Material recognition and the association of related properties can be helpful in causal decision-making \cite{Su22}. For example, an industrial robot can apply the optimum force for gripping an inelastic object if it knows the approximate material, mass, roughness, and size of this object. All this information relates to the physical density and material property.  

Machine learning offers solutions for material recognition \cite{Su22}. Databases such as Flickr \cite{Li10} are able to recognize different materials, as shown in Tab.~\ref{fig:FlickrDatabase}. 

\begin{table}[ht!]
    \centering
    \begin{tabular}{c c c c c}
       \hspace*{-0.2cm} \includegraphics[scale=0.2]{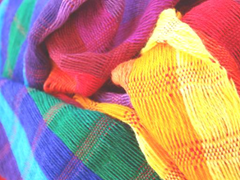}     & \hspace*{-0.2cm}
        \includegraphics[scale=0.2]{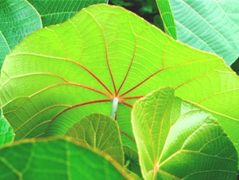}   & \hspace*{-0.2cm}
        \includegraphics[scale=0.2]{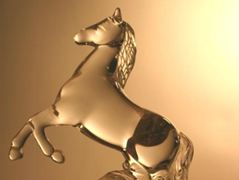}       & \hspace*{-0.2cm}
        \includegraphics[scale=0.2]{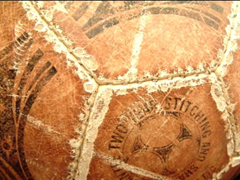}   &  \hspace*{-0.2cm}
        \includegraphics[scale=0.2]{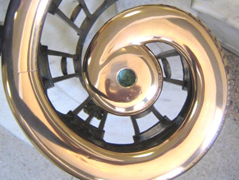}\\
        Fabric &
        Foliage &
        Glass &
        Leather &
        Metal \\
    \hspace*{-0.2cm}   \includegraphics[scale=0.2]{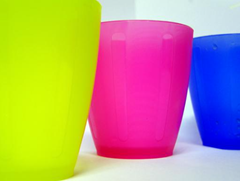}     &  \hspace*{-0.2cm}
      \includegraphics[scale=0.2]{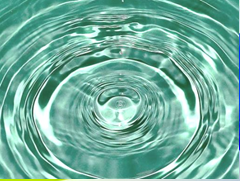}   & \hspace*{-0.2cm}
      \includegraphics[scale=0.2]{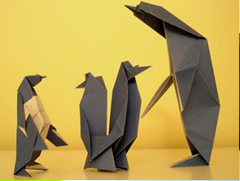}       & \hspace*{-0.2cm}
      \includegraphics[scale=0.2]{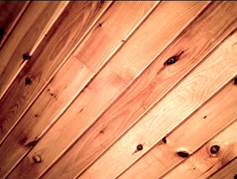}   &  \hspace*{-0.2cm}
       \includegraphics[scale=0.2]{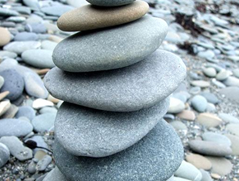}     \\  
               Plastic &
        Water &
        Paper &
        Wood &
        Stone \vspace*{0.2cm} \\

    \end{tabular}
    \caption{\textbf{State-of-the-Art Material Database \cite{Li10}.} Illustration of ten example material categories of Flickr database. The lighting conditions, compositions, colors, textures, surface shapes, material subtypes, and object associations were considered by an image diversity of 100 pictures, 50 close-ups, and 50 normal views in each category~\cite{FMD.24}. }
    \label{fig:FlickrDatabase}
\end{table}

These patterns are recognized by visual features and stored in the model through a learning process, using sophisticated AI algorithms for the recognition of objects in 2D images, such as YOLOv3 \cite{Red18}, Faster Region-Based Convolutional Neural Networks (R-CNN) \cite{Ren17}, and Multi-Scale Convolutional Neural Network (MSCNN) \cite{cai16}. However, the previously trained 2D object recognition is limited to specific object classes and visual interference effects of images \cite{MM22}. To collect ambient information and make accurate decisions with high confidence, the AI usually needs to be trained extensively with vast sets of data. Additionally, this material information has usually no connection with the associated physical properties. 

Based on the challenges of accurate processing and linking causal information, we present an approach that enables the assignment of physical properties in objects based on a 2D image by using machine learning and pattern recognition. The object is extracted and scaled into triangles to estimate the volume. We derive the associated materials from a database and ultimately calculate the density as a physical quality.

% R1: Was sind die Vorteile/der Mehrwert dieser contributions?
AI-based recognition of density and volume provides a solid foundation for the extraction of additional information from the environment. As an example, object-related forces can be calculated based on equation-specific coefficients, constants, and acquired sensor data including density. This expands the information content in a visual scene. Especially in road traffic, this can be an additional aid to improve the perception of autonomous vehicles.   

This paper describes a proof of concept for the implementation of AI-based density recognition. Our work comprises the following contributions:
\begin{itemize}
    \item Neural-specific object and texture detection based on object classification
    \item Concept of AI-based density recognition
    \item Analysis of recognized object density and material composition
\end{itemize}

Our evaluation reveals the feasibility and transferability of AI-based density recognition. For our empirical examination, we use synthetically generated data from the Unreal Engine.

The paper is organized according to a fixed structure consisting of related work, concept, methodology of AI-based object and texture detection, evaluation, conclusion, as well as future work.

%%%%%%%%%%%%%%%%%%%%%%%%%%%%%%%%%%%%%%%%%%%%%%%%%%%%%%%%%%%%%%%%%%%%%%%%%%%%%%%%%%%%%%%%%%%%%%%%%%

\section{Modern recognition}

% Einleitung mit Datenbank Flickr
This section presents recognition models and existing approaches for physical property recognition. The basic idea involves material recognition, which gives rise to entire databases such as Flickr \cite{Li10} which assigns materials based on visual appearance.

% ToDo:
% Flickr - Ansatz Material Erkennung/ Grundprinzip
% Urpsrung: https://people.csail.mit.edu/celiu/CVPR2010/maltRecogCVPR10.pdf
Liu et al. \cite{Li10} describe that the visual appearance of a surface depends on illumination conditions, geometric structure of surfaces at different spatial scales, and reflectance properties. Thereby, the reflectance properties of the surface are often characterized by features with a bidirectional reflectance distribution function (BRDF) \cite{N65}. In this context, material recognition can employ the recognition of colors and textures, micro-textures, outline shapes, or reflectance-based features \cite{Li10} by SIFT algorithms. This algorithm can recognize contours based on corners and edges \cite{K.16}.
% Gleichungsspezifische Zuordnung der Merkmale 

Standard k-means algorithms Eq.~\ref{eq:kMeansAlgorithm} are used to cluster instances of each feature~\cite{Li10} in order to assign the image-specific materials $M$ to the respective words.

\vspace*{-0.3cm}

\begin{equation}\label{eq:kMeansAlgorithm}
    J = \sum^{k}_\mathrm{i=1} \sum_\mathrm{x_\mathrm{j} \in S_\mathrm{i}} ||x_\mathrm{j} - \mu_\mathrm{i} ||^{2}
\end{equation}

\vspace*{-0.1cm}

$S_\mathrm{i}$ describes the cluster in Eq.~\ref{eq:kMeansAlgorithm}, which is determined from data points $x_\mathrm{j}$ and centroids $\mu_\mathrm{i}$ on the basis variance minimization~\cite{ES.00} and squared Euclidean distance $||x_\mathrm{j} - \mu_\mathrm{i} ||^{2}$. The random mean value $k$ is determined in the visual data set $m_\mathrm{i}$, ..., $m_\mathrm{k}$. Each data object is assigned to the cluster with the lowest variance for all $l = [1, ... , k]$, shown in Eq.~\ref{eq:ClusterAssignment}. 

\vspace*{-0.3cm}

% Lloyd-Algorithmus
\begin{equation}\label{eq:ClusterAssignment}
   S_\mathrm{i} = \{ x_\mathrm{j} : || x_\mathrm{j} - m_\mathrm{i}||^{2} \leq || x_\mathrm{j} - m_\mathrm{l} ||^{2} \}
\end{equation}

%Numerous methods have been developed to solve recognition problems by analyzing environmental events. An important example of this approach is machine learning, which uses static methods to continuously learn specifically from experience.The machine learning models are designed to map input variables to discrete output  variables. This process of predicting the labeling of given data is called classification. Regression refers to distinguishing between more than two categories. Based on the training, the data is divided into different classes \cite{Meh19}. 

% Quelle: https://towardsdatascience.com/how-to-define-a-neural-network-as-a-mathematical-function-f7b820cde3f

\begin{figure}[ht!]
    \centering
    \includegraphics[scale=0.5]{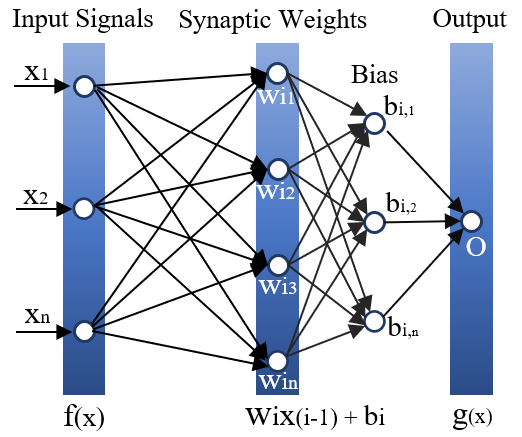}
    \caption{\textbf{Schematic Neural Network, adapted from~\cite{YYA.15}.} The input signals $f(i,x):$[$x_\mathrm{1}$, .. , $x_\mathrm{n}$] are weighted using the weighting factor $w_\mathrm{i}$ before they reach the main part of the neuron. A bias $b_\mathrm{i}$ is included as a threshold value which must first be exceeded to generate the output signal $O$.   }
    \label{fig:SchemeNeuralNetwork}
\end{figure}

Machine learning can be used as a static method to learn continuously and specifically from experiences. Thereby, the training-based data is divided into different classes. The classification $S_\mathrm{i}$ permits the mapping of input variables $f_\mathrm{i,x}: \mathbb{R}^{n_\mathrm{i}} \rightarrow \mathbb{R}^{n_\mathrm{i-1}}$ to discrete output variables $O$. In this view, the regression $g_\mathrm{(x)}$ refers to a distinction between more than two categories~\cite{Meh19}. Fig.~\ref{fig:SchemeNeuralNetwork} summarises the related components of neural networks.

Maschine learning techniques are broadly divided into two processes: feature extraction and classification. %The process of feature extraction and classification can be performed by a neural network.
Feature extraction involves the identification and correlation of patterns with large datasets suitable for modeling. Thereby, a feature refers to a property derived from raw data input intending to provide a suitable representation \cite{Jan21}.
In both cases, numerous processing nodes of neural networks are tightly interconnected and further layered in organized nodes to perform complex calculations \cite{SSBD14}. 

The overall quality of learning-based feature extraction depends on the task area and associated data set. In this respect, the data must contain a high density of information. To differentiate between classes, the algorithms try to recognize existing patterns of colors, shapes, textures or pixel values within this information \cite{Ipp19}. 
%To better differentiate between the classes, the algorithms attempt to recognize specific patterns within this information. These patterns can contain features such as colors, shapes, textures, or pixel values \cite{Ipp19}.

%Girshick et al.  describe an approach by applying high-capacity convolutional neural networks to bottom-up region proposals to localize and segment them. 

Girshick et al.~\cite{Gir15} describe an approach in which high-capacity convolutional neural networks are applied to bottom-up region proposals for localization and segmentation. They introduce a paradigm for training large CNNs when labeled training data is scarce. They detail how pre-training the network with supervision on an auxiliary task with abundant data (image classification) and then fine-tuning the network on the target task where data is scarce (recognition) can improve overall efficiency. This approach is similar to R-CNN. In fast R-CNN, the CNN is first fed with the input image to generate a convolutional feature map. Subsequently, the selective search is performed and the region suggestions are warped into squares. Those region suggestions are called Regions of Interest (RoI) and refer to a subset of the original image \cite{Kem20}. By using RoI pooling layers, each region that has been proposed and which may have different sizes, is reshaped into fixed size so that it can be fed into a fully connected layer. On the output, a softmax layer is used to predict the class of the proposed region and the values of the bounding box \cite{Mic15}. This method produces results faster since it calculates the CNN features only once per image and not two thousand times as with the R-CNN method. %However, other models even surpass R-CNN in speed and accuracy.

% Funktionsprinzip
Sean Bell et al.~\cite{BUSB15} suggest in a direct comparison between three different CNN models (AlexNet, VGG-16, GoogleNet) that material recognition and segmentation of everyday images which are based on Materials in Context Database (MINC) is possible with a probability of 82.2 $\%$ (AlexNet) to 85.9 $\%$ (GoogleNet). 
%Sean Bell et al. \cite{BUSB15} provide a data set as well as three different CNN models that are used to recognize materials. The data set distinguishes between 23 different classes with three million samples. A smaller data set with 1000 images per label, the Materials in Context Database (MINC) serves as the basis for material recognition and segmentation of everyday images in their paper. Their evaluation found that AlexNet with 82.2 $\%$, VGG-16 with 85.6 $\%$, and GoogleNet with 85.9 $\%$ provided the best results. %The authors also presented a concept for using material segmentation for detection. Of course there are other approaches to recognize specific patterns for the detection of materials. 
%Shukla et al.~\cite{SKK+21} evaluated the accuracy of CNN classifiers for material detection against deep-learning classifiers. They found that the proposed CNN leads to better recognition accuracy than deep-learning methods. 

Shukla et al ~\cite{SKK+21} evaluated the accuracy between CNN classifiers for material recognition and deep learning classifiers. They found that CNN classifiers have better and faster recognition accuracy since the existing probability level allows the classifier to recognize materials with higher accuracy.

Various previous works~ \cite{cai16,mue18,Red18,Ren17} refer to AI algorithms such as YOLOv3, Faster R-CNN and MSCNN that use a class-verifying diversity of object propositions for object recognition. 

%Prevalent AI algorithms such as YOLOv3, Faster R-CNN, and MSCNN use a class-verifying diversity of object propositions for object recognition \cite{cai16,mue18,Red18,Ren17}.
% YOLO architecture can implement tasks in the areas of general and oriented recognition, instance segmentation, pose, key points, as well as classification \cite{yolov8_ultralytics}. YOLO's computationally intensive operations required for machine learning are in line with its performance. 

Modern Architectures such as YOLO have been continuously improved to perform tasks in the areas of general and oriented recognition, instance segmentation, pose, key points and classification~ \cite{yolov8_ultralytics}. A DarkNet-19 model architecture (YOLOv2) was expanded to a more complex backbone model DarkNet-53 (YOLOv3) in which features on three different scales can be recognized~\cite{Dag21}. Although the implementation of such new functions and targeted optimizations reduce latency times, they often require computationally intensive operations that demand considerable computing power.

Ren et al. \cite{Ren17} found a solution to an issue of R-CNN caused by selective search. Selective search is a rigid algorithm that is unable to improve or learn, which can lead to poor suggestions for candidate regions. They developed the Faster-R-CNN algorithm, replacing selective search with a separate network, the Region Proposal Network (RPN), to predict region proposals. The RPN takes an image as input and outputs a series of rectangular object proposals, each with a class prediction and a confidence value. The network can be trained throughout by backpropagation, where the gradient of the loss function is calculated taking into account the weights of the network for a single input-output example \cite{RHW86}.

Wu et al. \cite{JJH+16} teach a computational vision system to understand physical relationships with the help of unlabelled videos. They address specific physical scenarios and distinguish between two groups of physical properties: The first inherits the intrinsic physical properties of objects such as volume, material, and mass. The other group is the descriptive physical properties, which
describe the scene and are determined by the first group. These include, but are not limited to, the speed of the objects, the distance they travel, or whether they fall into water. The presented model uses CNN to learn the object properties exclusively from unlabelled data. This approach provides serviceable results on a physical data set. 

% Zusammenfassung der vorherrigen arbeiten und überleitung zum Konzept

%\newpage

\section{Density Recognition}

%\begin{figure*}[ht!]{}
 %   \centering
    %\hspace*{0cm} \includegraphics[width=0.9\textwidth]{Pictures/ConceptV3.png}   
  %  \caption{\textbf{Fundamental Pipeline of AI-based Density Recognition}: }
   % \label{fig:PipelineOfDensityRecognition}
% \end{figure*}

\begin{figure}[htb!]
    \centering
    
\tikzset{every picture/.style={line width=0.75pt}} %set default line width to 0.75pt       
\begin{tikzpicture}[x=0.75pt,y=0.75pt,yscale=-1,xscale=1]
%uncomment if require: \path (0,4310); %set diagram left start at 0, and has height of 4310

%Shape: Parallelogram [id:dp15820942331471666] 
\draw   (207.71,2741.32) -- (164.36,2764.15) -- (164.29,2718.98) -- (207.64,2696.14) -- cycle ;
%Shape: Parallelogram [id:dp11567106994979981] 
\draw   (284.71,2740.32) -- (241.36,2763.15) -- (241.29,2717.98) -- (284.64,2695.14) -- cycle ;
%Straight Lines [id:da6784068160353394] 
\draw    (213.33,2726.15) -- (231.33,2726.15) ;
\draw [shift={(234.33,2726.15)}, rotate = 180] [fill={rgb, 255:red, 0; green, 0; blue, 0 }  ][line width=0.08]  [draw opacity=0] (10.72,-5.15) -- (0,0) -- (10.72,5.15) -- (7.12,0) -- cycle    ;
%Shape: Parallelogram [id:dp5249629666397748] 
\draw   (356.71,2738.32) -- (313.36,2761.15) -- (313.29,2715.98) -- (356.64,2693.14) -- cycle ;
%Straight Lines [id:da08679704482343586] 
\draw    (290.33,2726.15) -- (308.33,2726.15) ;
\draw [shift={(311.33,2726.15)}, rotate = 180] [fill={rgb, 255:red, 0; green, 0; blue, 0 }  ][line width=0.08]  [draw opacity=0] (10.72,-5.15) -- (0,0) -- (10.72,5.15) -- (7.12,0) -- cycle    ;
%Shape: Parallelogram [id:dp986211613556887] 
\draw  [color={rgb, 255:red, 126; green, 211; blue, 33 }  ,draw opacity=1 ] (266.54,2735.52) -- (246.36,2746.15) -- (246.15,2724.78) -- (266.33,2714.15) -- cycle ;
%Shape: Parallelogram [id:dp43321453263372556] 
\draw  [color={rgb, 255:red, 126; green, 211; blue, 33 }  ,draw opacity=1 ] (282.39,2734.89) -- (262.21,2745.52) -- (262,2724.15) -- (282.19,2713.51) -- cycle ;
%Shape: Parallelogram [id:dp5724023340105833] 
\draw   (232.71,2776.32) -- (189.36,2799.15) -- (189.29,2753.98) -- (232.64,2731.14) -- cycle ;
%Shape: Parallelogram [id:dp8378760604865108] 
\draw  [color={rgb, 255:red, 126; green, 211; blue, 33 }  ,draw opacity=1 ] (225.24,2773.6) -- (210.21,2781.52) -- (210.3,2766.07) -- (225.33,2758.15) -- cycle ;
%Straight Lines [id:da31289497097597674] 
\draw    (174.33,2747.15) -- (184.67,2762.65) ;
\draw [shift={(186.33,2765.15)}, rotate = 236.31] [fill={rgb, 255:red, 0; green, 0; blue, 0 }  ][line width=0.08]  [draw opacity=0] (10.72,-5.15) -- (0,0) -- (10.72,5.15) -- (7.12,0) -- cycle    ;
%Shape: Parallelogram [id:dp8520379651079748] 
\draw  [color={rgb, 255:red, 126; green, 211; blue, 33 }  ,draw opacity=1 ] (210.24,2773.6) -- (195.21,2781.52) -- (195.3,2766.07) -- (210.33,2758.15) -- cycle ;
%Shape: Parallelogram [id:dp54094682993474] 
\draw  [color={rgb, 255:red, 126; green, 211; blue, 33 }  ,draw opacity=1 ] (349.24,2734.6) -- (334.21,2742.52) -- (334.3,2727.07) -- (349.33,2719.15) -- cycle ;
%Shape: Parallelogram [id:dp2731615961417935] 
\draw  [color={rgb, 255:red, 126; green, 211; blue, 33 }  ,draw opacity=1 ] (334.24,2734.6) -- (319.21,2742.52) -- (319.3,2727.07) -- (334.33,2719.15) -- cycle ;
%Straight Lines [id:da7524929997309406] 
\draw    (359.33,2725.15) -- (375.33,2725.15) -- (390.26,2743.33) ;
\draw [shift={(392.17,2745.65)}, rotate = 230.61] [fill={rgb, 255:red, 0; green, 0; blue, 0 }  ][line width=0.08]  [draw opacity=0] (10.72,-5.15) -- (0,0) -- (10.72,5.15) -- (7.12,0) -- cycle    ;
%Straight Lines [id:da06517735063687424] 
\draw    (402,2751.15) -- (424.33,2751.15) ;
\draw [shift={(427.33,2751.15)}, rotate = 180] [fill={rgb, 255:red, 0; green, 0; blue, 0 }  ][line width=0.08]  [draw opacity=0] (10.72,-5.15) -- (0,0) -- (10.72,5.15) -- (7.12,0) -- cycle    ;
%Straight Lines [id:da8472865672023842] 
\draw    (305.33,2778.15) -- (354.1,2778.15) -- (386.79,2757.73) ;
\draw [shift={(389.33,2756.15)}, rotate = 148.02] [fill={rgb, 255:red, 0; green, 0; blue, 0 }  ][line width=0.08]  [draw opacity=0] (10.72,-5.15) -- (0,0) -- (10.72,5.15) -- (7.12,0) -- cycle    ;
%Shape: Circle [id:dp4221433453601633] 
\draw  [fill={rgb, 255:red, 74; green, 74; blue, 74 }  ,fill opacity=0.34 ] (390.17,2751.65) .. controls (390.17,2748.88) and (392.41,2746.65) .. (395.17,2746.65) .. controls (397.93,2746.65) and (400.17,2748.88) .. (400.17,2751.65) .. controls (400.17,2754.41) and (397.93,2756.65) .. (395.17,2756.65) .. controls (392.41,2756.65) and (390.17,2754.41) .. (390.17,2751.65) -- cycle ;
%Shape: Cube [id:dp9246830985194128] 
\draw  [color={rgb, 255:red, 0; green, 0; blue, 0 }  ,draw opacity=1 ][fill={rgb, 255:red, 0; green, 0; blue, 0 }  ,fill opacity=0.1 ] (285.32,2772.41) -- (298.31,2761.77) -- (302.88,2762.98) -- (303.03,2776.89) -- (290.04,2787.52) -- (285.48,2786.32) -- cycle ; \draw  [color={rgb, 255:red, 0; green, 0; blue, 0 }  ,draw opacity=1 ] (302.88,2762.98) -- (289.89,2773.61) -- (285.32,2772.41) ; \draw  [color={rgb, 255:red, 0; green, 0; blue, 0 }  ,draw opacity=1 ] (289.89,2773.61) -- (290.04,2787.52) ;
%Straight Lines [id:da28047932494625627] 
\draw    (236.33,2779.15) -- (280.33,2779.15) ;
\draw [shift={(283.33,2779.15)}, rotate = 180] [fill={rgb, 255:red, 0; green, 0; blue, 0 }  ][line width=0.08]  [draw opacity=0] (10.72,-5.15) -- (0,0) -- (10.72,5.15) -- (7.12,0) -- cycle    ;

% gestrichelte Linien Zuweisung 
%Straight Lines [id:da4488327066272695] 
\draw [color={rgb, 255:red, 0; green, 0; blue, 0 }  ,draw opacity=0.44 ] [dash pattern={on 4.5pt off 4.5pt}]  (243.33,2739.15) -- (211.33,2758.15) ;
%Straight Lines [id:da23815650612275818] 
\draw [color={rgb, 255:red, 0; green, 0; blue, 0 }  ,draw opacity=0.44 ] [dash pattern={on 4.5pt off 4.5pt}]  (262.21,2745.52) -- (230.21,2764.52) ;

%Straight Lines [id:da8597503920922551] 
\draw [color={rgb, 255:red, 155; green, 155; blue, 155 }  ,draw opacity=1 ] [dash pattern={on 4.5pt off 4.5pt}]  (291.3,2761.07) -- (291.33,2748.15) -- (310.33,2739.15) ;

% Text Node
\draw (163,2677.15) node [anchor=north west][inner sep=0.75pt]   [align=left] {\textbf{Image}};
% Text Node
\draw (217,2677.15) node [anchor=north west][inner sep=0.75pt]   [align=left] {\textbf{Object Detection}};
% Text Node
\draw (328,2677.15) node [anchor=north west][inner sep=0.75pt]   [align=left] {\textbf{Mesh Assignment}};
% Text Node
\draw (162,2803) node [anchor=north west][inner sep=0.75pt]   [align=left] {\textbf{Texture Detection}};
% Text Node
\draw (251.64,2762) node [anchor=north west][inner sep=0.75pt]   [align=left] {$\sum M$};
% Text Node
\draw (360,2708) node [anchor=north west][inner sep=0.75pt]   [align=left] {$\sum V$};
% Text Node
\draw (334.33,2723) node [anchor=north west][inner sep=0.75pt]   [align=left] {$V_\mathrm{n}$};
% Text Node
\draw (319,2725) node [anchor=north west][inner sep=0.75pt]   [align=left] {$V_\mathrm{1}$};
% Text Node
\draw (246,2725) node [anchor=north west][inner sep=0.75pt]   [align=left] {$b_\mathrm{1}$};
% Text Node
\draw (265,2722) node [anchor=north west][inner sep=0.75pt]   [align=left] {$b_\mathrm{n}$};
% Text Node
\draw (190,2764) node [anchor=north west][inner sep=0.75pt]   [align=left] {$M_\mathrm{1}$};
% Text Node
\draw (209.33,2760) node [anchor=north west][inner sep=0.75pt]   [align=left] {$M_\mathrm{n}$};
% Text Node
\draw (398,2732) node [anchor=north west][inner sep=0.75pt]   [align=left] {$ \sum m $};
% Text Node
\draw (321.64,2762) node [anchor=north west][inner sep=0.75pt]   [align=left] {$\sum \varrho $};
% Text Node
\draw (270,2790) node [anchor=north west][inner sep=0.75pt]   [align=left] {\textbf{Density Assignment}};
% Text Node
\draw (401,2755.15) node [anchor=north west][inner sep=0.75pt]   [align=left] {\textbf{Mass}};
\end{tikzpicture}
    \caption{\textbf{Pipeline of AI-based density recognition.} The pipeline includes the detection of objects and their textures as well as the assignment of density and meshes to calculate physical quantities like object masses.}
    \label{fig:PipelineOfAIbasedDensityRecognition}
\end{figure}
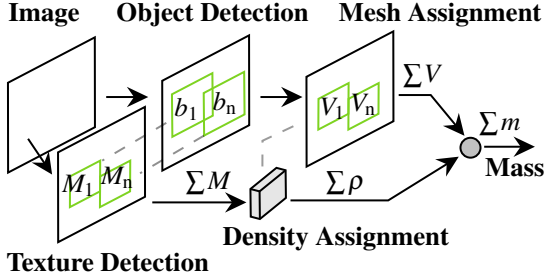

This chapter describes the concept of density and material recognition in order to calculate physical properties like masses. Fig.~\ref{fig:PipelineOfAIbasedDensityRecognition} illustrates the fundamental pipeline of AI-based density recognition.

Building on identifying specific features from 2D images of previous work, we combine object detection with the assignment and calculation of physical properties. The image data is first analyzed by using a neural network. Salient objects can be identified and classified texturally. Object areas are identified as $b_\mathrm{1}, ..., b_\mathrm{n}$.  Within the bounding boxes, possible materials can be attributed based on Flickr Material Database (FMD) and Materials in Context Database (MINC). In this process, these materials $M$ will be assigned a specific literary density $\rho$ where we consider further pattern properties like image colors, shapes, textures and pixels. Since it is necessary to determine the volume $V$ of the respective objects, the process calculates their corresponding meshes. The physical properties like mass $M$ can be derived from the extensive information.

%\begin{figure}[ht!]
%    \centering
%    \includegraphics[scale=0.5]{picture/Input.png}
%    \caption{\textbf{Input Data Set}}
%    \label{fig:InputData}
%\end{figure}

\subsection{Object Detection}

%Numerous models enable us to identify the desired objects with their key points. %, as shown in Fig.~\ref{fig:ClassifiedObjectDetection}. 

\begin{figure}[ht!]
    \centering
    \includegraphics[scale=0.32]{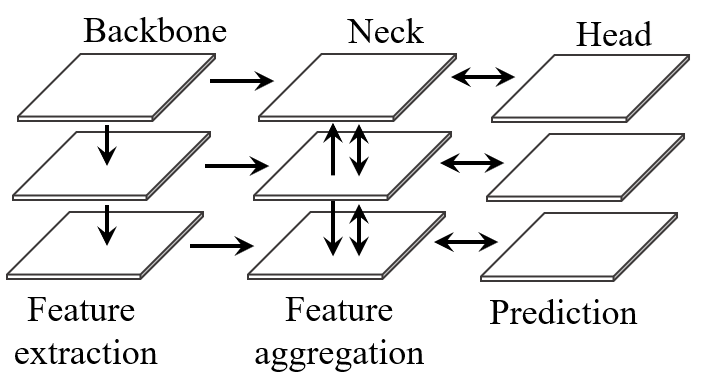}
    \caption{\textbf{YOLO Detectors used in this Work.} The backbone extracts important features from the image at different scales. The neck concatenates the semantic information from different layers of the backbone network and transmits it as input to the head. The head applies the refined features for predictive object recognition. }
    \label{fig:YOLOprincipleV1}
\end{figure}

In order to detect objects, we use the convolutional network of 
YOLOv4~\cite{yolov8_ultralytics}. We extract features from the input images and calculate them into feature maps. As part of the YOLO architecture, we use backbone, neck, and head detectors, as shown in Fig.~\ref{fig:YOLOprincipleV1}. 

\vspace*{-0.3cm}

YOLOv4 contains a pre-trained convolutional neural network such as VGG16 or CSPDarkNet53 as a backbone which is based on SPP-Modul (Spatial Pyramid Pooling) and PAN (Path Aggregation Network). As part of the prediction, the head processes aggregated features and predicts the bounding boxes, objecthood, and classification values.

Our model is trained on a MS COCO dataset, which contains over 80 different classes and 1.5 million object instances in 200 thousand labeled images.

% Merkmalspyramidennetzwerk (FPN) Thereby, the backbone is pretrained on ImageNet. 

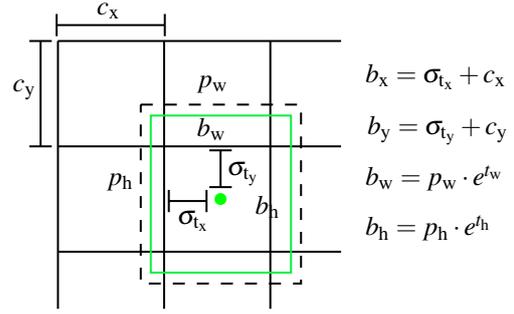
\begin{figure}[ht!]
    \centering

\tikzset{every picture/.style={line width=0.75pt}} %set default line width to 0.75pt        

\begin{tikzpicture}[x=0.75pt,y=0.75pt,yscale=-1,xscale=1]
%uncomment if require: \path (0,4310); %set diagram left start at 0, and has height of 4310

%Shape: Grid [id:dp7416061424631915] 
\draw  [draw opacity=0] (100,2905.81) -- (241.33,2905.81) -- (241.33,3040.48) -- (100,3040.48) -- cycle ; \draw   (100,2905.81) -- (100,3040.48)(153,2905.81) -- (153,3040.48)(206,2905.81) -- (206,3040.48) ; \draw   (100,2905.81) -- (241.33,2905.81)(100,2958.81) -- (241.33,2958.81)(100,3011.81) -- (241.33,3011.81) ; \draw    ;
%Shape: Rectangle [id:dp9460076338295484] 
\draw  [color={rgb, 255:red, 12; green, 239; blue, 64 }  ,draw opacity=1 ] (146.33,2943.31) -- (216.33,2943.31) -- (216.33,3022.31) -- (146.33,3022.31) -- cycle ;
%Straight Lines [id:da9804510705030351] 
\draw    (99.33,2897.81) -- (153.33,2897.81) ;
\draw [shift={(153.33,2897.81)}, rotate = 180] [color={rgb, 255:red, 0; green, 0; blue, 0 }  ][line width=0.75]    (0,5.59) -- (0,-5.59)   ;
\draw [shift={(99.33,2897.81)}, rotate = 180] [color={rgb, 255:red, 0; green, 0; blue, 0 }  ][line width=0.75]    (0,5.59) -- (0,-5.59)   ;
%Straight Lines [id:da2678913451162428] 
\draw    (91.33,2905.81) -- (91.33,2958.81) ;
\draw [shift={(91.33,2958.81)}, rotate = 270] [color={rgb, 255:red, 0; green, 0; blue, 0 }  ][line width=0.75]    (0,5.59) -- (0,-5.59)   ;
\draw [shift={(91.33,2905.81)}, rotate = 270] [color={rgb, 255:red, 0; green, 0; blue, 0 }  ][line width=0.75]    (0,5.59) -- (0,-5.59)   ;
%Shape: Rectangle [id:dp8393564034976064] 
\draw  [dash pattern={on 4.5pt off 4.5pt}] (141.33,2937.81) -- (221.33,2937.81) -- (221.33,3027.81) -- (141.33,3027.81) -- cycle ;
%Shape: Circle [id:dp4814363596315876] 
\draw  [draw opacity=0][fill={rgb, 255:red, 4; green, 255; blue, 0 }  ,fill opacity=1 ] (178.17,2985.31) .. controls (178.17,2983.66) and (179.51,2982.31) .. (181.17,2982.31) .. controls (182.82,2982.31) and (184.17,2983.66) .. (184.17,2985.31) .. controls (184.17,2986.97) and (182.82,2988.31) .. (181.17,2988.31) .. controls (179.51,2988.31) and (178.17,2986.97) .. (178.17,2985.31) -- cycle ;
%Straight Lines [id:da8067106999279685] 
\draw    (181.17,2960.81) -- (181.17,2979.31) ;
\draw [shift={(181.17,2979.31)}, rotate = 270] [color={rgb, 255:red, 0; green, 0; blue, 0 }  ][line width=0.75]    (0,5.59) -- (0,-5.59)   ;
\draw [shift={(181.17,2960.81)}, rotate = 270] [color={rgb, 255:red, 0; green, 0; blue, 0 }  ][line width=0.75]    (0,5.59) -- (0,-5.59)   ;
%Straight Lines [id:da9151905993857534] 
\draw    (155.33,2986.31) -- (174.17,2986.31) ;
\draw [shift={(174.17,2986.31)}, rotate = 180] [color={rgb, 255:red, 0; green, 0; blue, 0 }  ][line width=0.75]    (0,5.59) -- (0,-5.59)   ;
\draw [shift={(155.33,2986.31)}, rotate = 180] [color={rgb, 255:red, 0; green, 0; blue, 0 }  ][line width=0.75]    (0,5.59) -- (0,-5.59)   ;

% Text Node
\draw (118,2885) node [anchor=north west][inner sep=0.75pt]   [align=left] {$c_\mathrm{x}$};
% Text Node
\draw (76,2924) node [anchor=north west][inner sep=0.75pt]   [align=left] {$c_\mathrm{y}$};
% Text Node
\draw (183,2965) node [anchor=north west][inner sep=0.75pt]   [align=left] {$\sigma_\mathrm{t_\mathrm{y}}$};
% Text Node
\draw (158,2990) node [anchor=north west][inner sep=0.75pt]   [align=left] {$\sigma_\mathrm{t_\mathrm{x}}$};
% Text Node
\draw (168,2922) node [anchor=north west][inner sep=0.75pt]   [align=left] {$p_\mathrm{w}$};
% Text Node
\draw (123,2971) node [anchor=north west][inner sep=0.75pt]   [align=left] {$p_\mathrm{h}$};
% Text Node
\draw (168,2943) node [anchor=north west][inner sep=0.75pt]   [align=left] {$b_\mathrm{w}$};
% Text Node
\draw (197,2982) node [anchor=north west][inner sep=0.75pt]   [align=left] {$b_\mathrm{h}$};
% Text Node
\draw (252,2915.81) node [anchor=north west][inner sep=0.75pt]   [align=left] {$b_\mathrm{x} = \sigma_\mathrm{t_\mathrm{x}} + c_\mathrm{x} $};
% Text Node
\draw (253,2941.81) node [anchor=north west][inner sep=0.75pt]   [align=left] {$b_\mathrm{y} = \sigma_\mathrm{t_\mathrm{y}} + c_\mathrm{y} $};
% Text Node
\draw (252,2966.81) node [anchor=north west][inner sep=0.75pt]   [align=left] {$b_\mathrm{w} = p_\mathrm{w} \cdot e^{t_\mathrm{w}} $};
% Text Node
\draw (252,2991.81) node [anchor=north west][inner sep=0.75pt]   [align=left] {$b_\mathrm{h} = p_\mathrm{h} \cdot e^{t_\mathrm{h}} $};

\end{tikzpicture}
    \caption{\textbf{Bounding Boxes with Dimension Priors and Location
Prediction, adapted from~\cite{RF.17}.} The center coordinates of the box can be calculated with the predicted values $t_\mathrm{x}, t_\mathrm{y}$ using a sigmoid function and offset by the location of grid cell $c_\mathrm{x}, c_\mathrm{y}$. The width and height of the final box are adjusted to the previous width $p_\mathrm{w}$ and height $p_\mathrm{h}$ and scaled by $e^{t_\mathrm{w}}$ and $e^{t_\mathrm{h}}$.}
    \label{fig:GridCellBoundingBox}
\end{figure}

%~\cite{TCERG23}

Recognition first divides the image into a grid of cells like shwon in the example in Fig.~\ref{fig:GridCellBoundingBox}. The number of cells depends on the size of the image. For example, With a size of $608 \times 608$ pixels, the cell size is usually $32 \times 32$ pixels. Our data set is divided into $19 \times 19$ cells. 

\begin{figure}[ht!]
    \centering

\tikzset{every picture/.style={line width=0.75pt}} %set default line width to 0.75pt        

\begin{tikzpicture}[x=0.75pt,y=0.75pt,yscale=-1,xscale=1]
%uncomment if require: \path (0,4310); %set diagram left start at 0, and has height of 4310

%Image [id:dp1632905573150918] 
\draw (575.33,123.97) node  {\includegraphics[width=85.5pt,height=62.25pt]{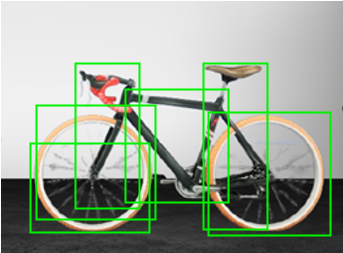}};
%Image [id:dp35415486526362017] 
\draw (698.83,123.97) node  {\includegraphics[width=84.75pt,height=62.25pt]{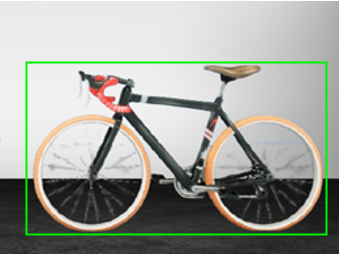}};
%Straight Lines [id:da48255694223010726] 
\draw    (555,169.64) -- (555,185.47) -- (715.86,185.47) -- (715.86,170.64) ;
\draw [shift={(715.86,167.64)}, rotate = 90] [fill={rgb, 255:red, 0; green, 0; blue, 0 }  ][line width=0.08]  [draw opacity=0] (10.72,-5.15) -- (0,0) -- (10.72,5.15) -- (7.12,0) -- cycle    ;

% Text Node
\draw (572,172) node [anchor=north west][inner sep=0.75pt]   [align=left] {Non-Max Suppression};
% Text Node
\draw (554,68) node [anchor=north west][inner sep=0.75pt]   [align=left] {\textbf{Before}};
% Text Node
\draw (686,68) node [anchor=north west][inner sep=0.75pt]   [align=left] {\textbf{After}};
\end{tikzpicture}
    
    \caption{\textbf{Effect of Non-Max Suppression (NMS).} The post-processing technique Non-maximum suppression reduces the number of overlapping bounding boxes. }
    \label{fig:NonMaximumSuppressionV1}
\end{figure}
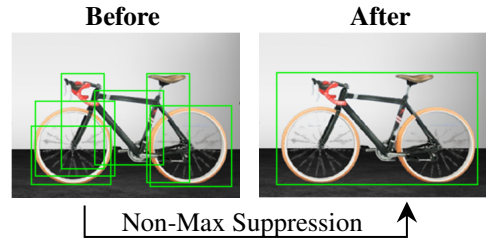

Each object is assigned to exactly one cell, containing the object's center point. Objectless cells are filtered out according to their low probability of all 80 classes. 
The use of Non-Max-Suppression, as shown in Fig.~\ref{fig:NonMaximumSuppressionV1}, eliminates unwanted bounding boxes so that only the most probable bounding box remains for each detected object.

% Non-Max Suppression

\iffalse
\begin{equation}
    IoU = \dfrac{A_\mathrm{Overlapping}}{A_\mathrm{Union}}
\end{equation}
\fi

The bounding boxes localize the position of objects in order to recognize possible textures and calculate the object-specific volumes. The area is adjusted to $224 \times 224$ pixels for the matching designation of the model input. 

\subsection{Texture Detection}

The object-specific bounding box detection enables the continuous application of a material detection model to the image area of the box. 
% Ergänzung zur konzept-relevanten Materialerkennung -> 
% Welche Methodik verwenden wir, was sind die wichtigen Gleichungen?
Inside the box, features such as color, SIFT, jet, micro-SIFT, micro-jet, curvature, edge-slice and edge-ribbon are combined and quantized into visual words by Bayesian framework~\cite{Li10}. 

Three different models are used, consisting of the MINC dataset. MINC comprises 23 different classes, each with 2500 images.  We measure the confidence score for all three models across MINC dataset and Flickr Material Database (FMD) for appropriate model selection (see also Tab.~\ref{tab:RecognitionOnFMD}). 

\begin{table}[ht!]
    \centering
    \begin{tabular}{r c c c }
     & & & \\ \hline
    & & & \vspace*{-0.3cm} \\ 
        \textbf{Classes} & \textbf{VGG16}  &   \textbf{GoogleNet} & \textbf{AlexNet} \vspace*{-0.3cm} \\ 
    & & & \\ \hline
    & & & \vspace*{-0.3cm} \\ 
    Fabric & (78~|~78)  & (69~|~78)  & (45~|~64)  \\
    Foliage & (71~|~95) & (68~|~95)  & (62~|~93)  \\ 
    Glass & (40~|~82) & (40~|~84)  & (27~|~78)  \\
    Leather & (29~|~88)  & (18~|~84) & (9~|~80)  \\
    Metal & (44~|~72) & (37~|~76)  & (33~|~69)  \\ 
    Paper & (41~|~90)  & (35~|~90) & (11~|~85)   \\
    Plastic & (78~|~75)  & (84~|~78)  & (74~|~68)  \\
    Stone & (78~|~89)  & (62~|~87) & (52~|~85)  \\  
    Water & (47~|~96)  & (43~|~94)  & (30~|~93)  \\  
    Wood & (36~|~74) & (25~|~78) & (13~|~71)  \vspace*{-0.3cm} \\
    & & & \\ \hline
     & & & \vspace*{-0.3cm} \\ 
    \end{tabular}
    \caption{\textbf{Recognition on (FMD | MINC)}. VGG16 slightly outperforms GoogleNet. AlexNet provides the lowest accuracy out of these three contrasted models.} %~[\%]
    \label{tab:RecognitionOnFMD}
\end{table}

%\begin{table}[ht!]
%    \centering
%    \begin{tabular}{r c c c }
%     & & & \\ \hline
%    & & & \vspace*{-0.3cm} \\ 
%        \textbf{Classes} & \textbf{VGG16}  &   \textbf{GoogleNet} & \textbf{AlexNet} \vspace*{-0.3cm} \\ 
%    & & & \\ \hline
%    & & & \vspace*{-0.3cm} \\ 
%    Fabric & 78~\% &  78~\% &  64~\% \\
%    Foliage & 95~\% & 95 ~\% & 93~\% \\ 
%    Glass & 82~\% &  84~\% & 78~\% \\
%    Leather & 88~\% &  84~\% & 80~\% \\
%    Metal & 72~\% &  76~\% & 69~\% \\ 
%    Paper & 90~\% &  90~\% & 85~\%  \\
%    Plastic & 75~\% & 78~\% & 68~\% \\
%    Stone & 89~\% & 87~\% & 85~\% \\  
%    Water & 96~\% & 94~\% & 93~\% \\  
%    Wood & 74~\% & 78~\% &  71~\% \vspace*{-0.3cm} \\
%    & & & \\ \hline
%     & & & \vspace*{-0.3cm} \\ 
%    \end{tabular}
%    \caption{\textbf{Recognition on MINC}}
%    \label{tab:RecognitionOnMINC}
%\end{table}

FMD consists of ten classes with 100 images of each. Tab.~\ref{tab:RecognitionOnFMD} shows the performance of the models for the FMD dataset. VGG16 provides optimum results with an overall accuracy of 86~\% at MINC  and 52~\% at FMD. The model's size, however, requires significant time for both training and recognition. Alternatively, GoogleNet offers a good replacement model with an accuracy of 86~\% for MINC and 47~\% for FMD. The use of so-called inception modules allows a shorter calculation time. This module replaces a sparse CNN with a normal dense construction since most activations in a deep network are zero values or redundant due to correlations. As a result, not all output channels are connected to the input channels, hence the reduced computing time \cite{IEE15}.

%\subsection{Mesh Assignment}
\subsection{Mesh and Density Assignment}

Through the accompanying object recognition, the next step is to assign a suitable 3D mesh. The mesh must have certain properties for the correct calculation of volume. For example, each triangle of the mesh must have corner points stored in a clockwise direction. The mesh must be complete without open areas and completely closed to prevent subsequent miscalculations. Eq.~\ref{eq:VolumeCalculation2} describes the surface calculation of each signed triangle with $V_\mathrm{i} \in V$ and $i \in [1,n]$:

\begin{equation}\label{eq:VolumeCalculation2}
    V_\mathrm{i}^{'} = \dfrac{1}{6}(-x_\mathrm{i,3}y_\mathrm{i,2}z_\mathrm{i,1} +x_\mathrm{i,2}y_\mathrm{i,3}z_\mathrm{i,1} + 
\end{equation}
\begin{equation*}\label{eq:VolumeCalculation1}
   x_\mathrm{i,2}y_\mathrm{i,3}z_\mathrm{i,1} + x_\mathrm{i,3}y_\mathrm{i,1}z_\mathrm{i,2} -  x_\mathrm{i,1}y_\mathrm{i,3}z_\mathrm{i,2}-
\end{equation*}
\begin{equation*}
-x_\mathrm{i,2}y_\mathrm{i,1}z_\mathrm{i,3} + x_\mathrm{i,1}y_\mathrm{i,2}z_\mathrm{i,3})
\end{equation*}

\begin{figure}[ht!]
    \centering
    \includegraphics[scale=0.5]{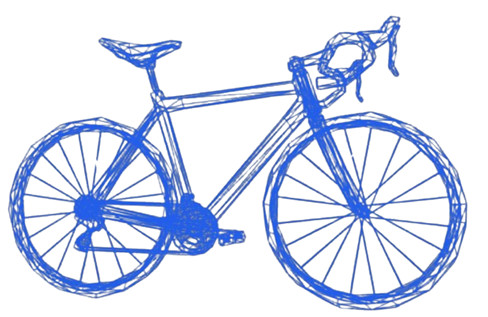}
    \caption{\textbf{Object Detection and Triangle Estimation.} Triangles: 5.022, Vertices: 4.159, UV Channels: 4 with approx. Size: 50 $\times$ 198 $\times$ 114.}
    \label{fig:ExampleMeshBicycle}
\end{figure}

We use i as the index for the triangles. $x_\mathrm{i,[1,2,3]}$, $y_\mathrm{i,[1,2,3]}$ and $z_\mathrm{i,[1,2,3]}$  are the coordinates of the vertices of triangle i.
Various shapes of objects are not considered in our analysis. Instead, each class is assigned to a specific 3D mesh. Fig.~\ref{fig:ExampleMeshBicycle} exemplifies the resulting 3D mesh calculated by Eq.~\ref{eq:VolumeCalculation2}.

After recognition, the applicable model is assigned. The object's mass is attributed based on the result of texture recognition. We simplify our analysis by assuming solid material for the respective model. A further database is created for the classification, which specifies the density of each recognizable material.

%\newpage

\subsection{Physical Properties}

% Konventionen in der Wärmeübertragung hängen von der Dichte ab. 

Physical quantities such as forces, friction, pressures, temperatures, air resistances, inertia moments, energies or material properties $n$ depend on density $\varrho$. The fundamental calculation of density $( \varrho = m / V )$ and the inclusion of further coefficients and constants offers possibilities of inferring different values.  

By recognizing the actual object and the possible material assignment, we can deduce the volume and mass of the object. The physical property is determined by iterating over each section since many objects are divided into sub-objects. The bicycle in Fig.~\ref{fig:ExampleMeshBicycle}, for example, includes the individual wheels, the handlebars, the saddle, and the frame. For each section, the signed volume is now calculated for each triangle and added to the total volume. Surfaces that point outwards contribute to the total volume. Surfaces that point inwards subtract from it. This leaves only the volume on the inside. The density and volume can then be multiplied to calculate the weight of the object.

\section{Experimental Setup}

Our approach utilizes Unreal Engine 4.27 and its high-fidelity rendering pipeline. The realistic rendering and lighting allow us to assume real test simulations as shown in Tab.~\ref{tab:SelectionofItems}. The neuronal training is based on realistic test images. 

\begin{table}[ht!]
    \centering
    \begin{tabular}{c c c c}
        
    \includegraphics[scale=0.2]{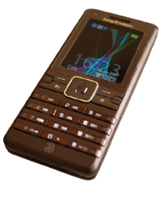}  &  
    \includegraphics[scale=0.2]{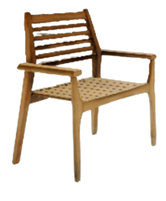} &  
    \includegraphics[scale=0.2]{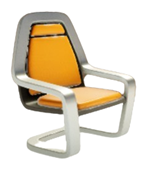} & 
    \includegraphics[scale=0.2]{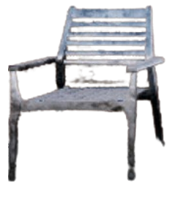} \\ 
   
    \hypertarget{Phone1}{Phone}   &  
    \hypertarget{Chair1}{Chair 1} &  
    \hypertarget{Chair2}{Chair 2} & 
    \hypertarget{Chair3}{Chair 3}  \\ 
            
    \includegraphics[scale=0.2]{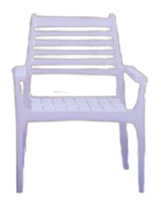}  &  
    \includegraphics[scale=0.25]{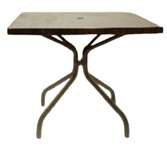}  &
    \includegraphics[scale=0.2]{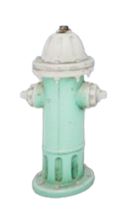} &
    \includegraphics[scale=0.2]{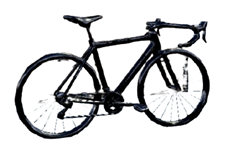}   \\
   
    \hypertarget{Chair4}{Chair 4} &
    \hypertarget{Table1}{Table 1} &
    \hypertarget{Hydrant}{Hydrant} &
    \hypertarget{Bicycle1}{Bicycle 1} \\
       
    \includegraphics[scale=0.2]{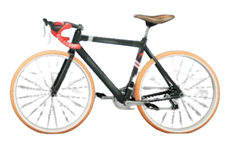} & 
    \includegraphics[scale=0.2]{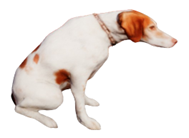} & 
    \includegraphics[scale=0.2]{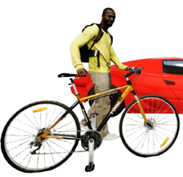} & 
    \includegraphics[scale=0.2]{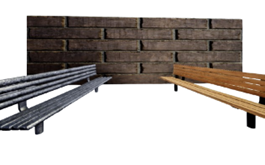}   \\ 
   
    \hypertarget{Bicycle2}{Bicycle 2} & 
    \hypertarget{Dog}{Dog} & 
    \hypertarget{Scene1}{Scene 1} & 
    \hypertarget{Scene2}{Scene 2}   \\       
    & & &  \\  
    \end{tabular}
    \caption{\textbf{Sample of Items used in our Experiment.} We used Unreal Engine 4.27 for its realistic rendering capabilities.}
    \label{tab:SelectionofItems}
\end{table}

The accuracy of the physical information is directly linked to the image quality. In our conceptual consideration, we use images with a size of $608 \times 608$ pixels and three assigned color channels. %, shown in Fig.~\ref{fig:InputData}.

Our computing hardware has an integrated Quad-Core Intel or AMD, 2.5 GHz, 8GB RAM, external GPU1 Nvidia GeForce GTX 1050 Ti, and onboard GPU0 of Intel HD Graphics 630. Cv2 is used for image processing and reading deep neural networks, and NumPy for mathematical functions. Mean subtraction calculates the average pixel intensity over all images of the used training set of all three color channels and subtracts these values from the channels of the input image. When using YOLOv4, channel swapping is also applied for the mean subtraction. Here the image is swapped in RGB order. In order to obtain one optimal bounding box for each object, non-max suppression is applied. To predict the material, we iterate over each object and use the corresponding image area as input to the texture model, similar to the object detection model. 

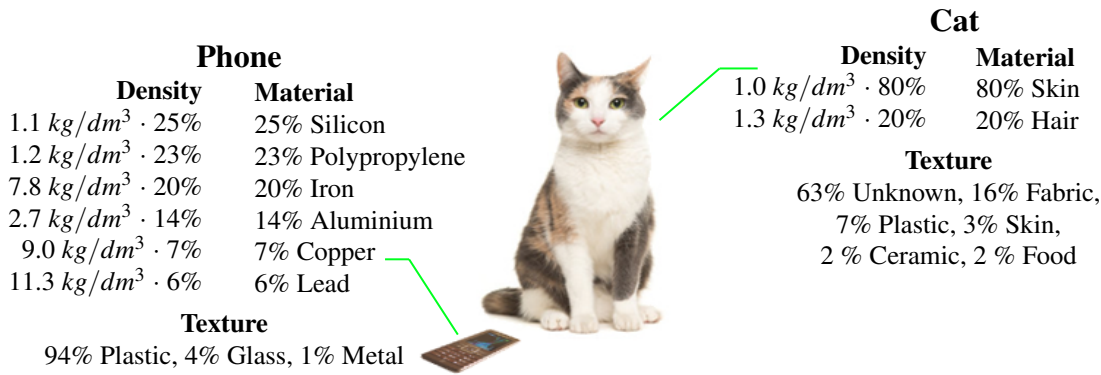
\begin{figure*}[!ht]{}

\tikzset{every picture/.style={line width=0.75pt}} %set default line width to 0.75pt        

\begin{tikzpicture}[x=0.75pt,y=0.75pt,yscale=-1,xscale=1]
%uncomment if require: \path (0,4310); %set diagram left start at 0, and has height of 4310

%Image [id:dp9468851858448322] 
\draw (376.33,179.07) node  {\includegraphics[width=102.5pt,height=127.6pt]{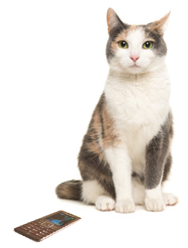}};
%Straight Lines [id:da8237500553428654] 
\draw [color={rgb, 255:red, 0; green, 255; blue, 31 }  ,draw opacity=1 ]   (311.67,203.14) -- (336,241) ;
%Straight Lines [id:da45414077227004035] 
\draw [color={rgb, 255:red, 0; green, 255; blue, 31 }  ,draw opacity=1 ]   (299.67,203.14) -- (311.67,203.14) ;
%Straight Lines [id:da8422402352290996] 
\draw [color={rgb, 255:red, 0; green, 255; blue, 31 }  ,draw opacity=1 ]   (436.67,133.14) -- (466.67,107.14) ;
%Straight Lines [id:da7452149520948512] 
\draw [color={rgb, 255:red, 0; green, 255; blue, 31 }  ,draw opacity=1 ]   (466.67,107.14) -- (485.67,107.14) ;

% Text Node
\draw (233,113) node [anchor=north west][inner sep=0.75pt]   [align=left] {\textbf{Material} \ \\
25$\%$ Silicon \\
23$\%$ Polypropylene \ \ \\
20$\%$ Iron \ \ \\
14$\%$ Aluminium\\
7$\%$ Copper \ \\
6$\%$ Lead };
% Text Node
\draw (90,112) node [anchor=north west][inner sep=0.75pt]   [align=left] {\begin{minipage}[lt]{87.77pt}\setlength\topsep{0pt}
\begin{flushright}
\textbf{Density}\\
1.1 $kg/dm^{3}$ $\cdot$ 25$\%$ \\ % 1.07
1.2  $kg/dm^{3}$ $\cdot$ 23$\%$ \\ 
7.8  $kg/dm^{3}$ $\cdot$ 20$\%$  \\
2.7  $kg/dm^{3}$ $\cdot$ 14$\%$ \\
9.0  $kg/dm^{3}$ $\cdot$ 7$\%$ \\ % 8.96 
11.3  $kg/dm^{3}$ $\cdot$ 6$\%$ 
\end{flushright}
\end{minipage}};
% Text Node
\draw (113,229) node [anchor=north west][inner sep=0.75pt]   [align=left] {\begin{minipage}[lt]{157.48pt}\setlength\topsep{0pt}
\begin{center}
\textbf{Texture}\\ 94$\%$ Plastic, 4$\%$ Glass, 1$\%$ Metal
\end{center}

\end{minipage}};
% Text Node

\draw (593,95) node [anchor=north west][inner sep=0.75pt]   [align=left] {\textbf{Material} \\
80$\%$ Skin \\
20$\%$ Hair};
% Text Node
\draw (438,94) node [anchor=north west][inner sep=0.75pt]   [align=left] {\begin{minipage}[lt]{97.96pt}\setlength\topsep{0pt}
\begin{flushright}
\textbf{Density} \\
1.0 $kg/dm^{3}$ $\cdot$ 80$\%$ \\
1.3 $kg/dm^{3}$ $\cdot$ 20$\%$
\end{flushright}

\end{minipage}};

% Text Node
\draw (484,147) node [anchor=north west][inner sep=0.75pt]   [align=left] {\begin{minipage}[lt]{143.32pt}\setlength\topsep{0pt}
\begin{center}
\textbf{Texture} \\ 63$\%$ Unknown, 16$\%$ Fabric, \\
7$\%$ Plastic,  3$\%$ Skin, \\
2 $\%$ Ceramic, 2 $\%$ Food
\end{center}

\end{minipage}};
% Text Node
\draw (204,94) node [anchor=north west][inner sep=0.75pt]   [align=left] {\textbf{\large Phone}};
% Text Node
\draw (570,76) node [anchor=north west][inner sep=0.75pt]   [align=left] {\textbf{\large Cat}};

\end{tikzpicture}

    \centering
    \caption{\textbf{Material and Density Composition of Recognized Objects.} The inner ingredients are frequently more complex and diverse than the exterior texture suggests. The illustration describes the composition of two example objects: A smartphone and cat. The density of occluded components can be estimated from the average composition of each object.}
    \label{fig:MaterialComposition}
\end{figure*}

%%%%%%%%%%%%%%%%%%%%%%%%%%%%%%%%%%%%%%%%%%%%%%%%%%%%%%%%%%%%%%%%%%%%%%%%%%%%%%%%%%%%%%%%%%%%%%%%%%
\section{Evaluation}

Our evaluation employs density recognition of several diverse objects. 

We used a Convolutional Neural Network to recognize the texture of the object and MINC for texture recognition. The model trained and used in our approach shows the highest accuracy when compared to other CNN architectures. 

In order to evaluate the functionality of our approach, we explored and tested select scenarios (see Tab.~\ref{tab:SelectionofItems}). We summarize our results in Tab.~\ref{tab:DetectabilityofMaterials}.

\begin{table}[ht!]
    \centering
    \begin{tabular}{r r l}
        &   &   \\ \hline
        &   &   \vspace*{-0.3cm} \\ 
        & \textbf{Material} & \textbf{Density} $[kg/dm^{3}]$ \\
        & \textbf{Type} & \textbf{(Literary | Measured) }   \vspace*{-0.3cm} \\
        &   &  \\ \hline
        &   &  \vspace*{-0.3cm} \\ 
        \textbf{\hyperlink{Phone1}{Phone}} &  Plastic & (4.0 | 1.2) $\pm$ 70~\% \\
        \textbf{Chair 1}  &  Wood & (0.7 | 0.7)  $\pm$ 4~\% \\
        \textbf{Chair 2}  &  Plastic & (4.8 | 1.2 ) $\pm$ 75~\% \\
        \textbf{Chair 3}  &  Metal & (7.9 | 8.0) $\pm$ 1~\% \\
        \textbf{Chair 4}  &  Metal & (0.9 | 1.2) $\pm$ 30~\% \\
        \textbf{Table 1}  &  Metal & (7.9 | 8.0) $\pm$ 1~\% \\
        \textbf{Hydrant}  &  Metal & (7.9 | 8.2) $\pm$ 14~\% \\
        \textbf{Bicycle 1}  &  Metal & (2.9 | 8.0) $\pm$ 180~\% \\
        \textbf{Bicycle 2}  &  Metal & (7.9 | 8.0) $\pm$ 2~\% \\
        \textbf{Bench 1}  &  Metal & (7.9 | 8.0) $\pm$ 1~\% \\
        \textbf{Bench 2}  &  Wood & (2.1 | 0.7) $\pm$ 66~\% \\
        \textbf{Dog}  &  Other & (1.1 | 1.0) $\pm$ 5.6~\% \\
        \textbf{Person}  &  Plastic & (1.1 | 1.0) $\pm$ 9~\% \\
        \textbf{Backback}  &  Fabric & (1.4 | 1.6) $\pm$ 16~\% \\
        \textbf{Car}  &  Plastic & (5.4 | 1.2) $\pm$ 77~\% \\
        &  &    \vspace*{-0.6cm} \\ 
        &  &   \\ \hline
        &  &    \vspace*{-0.3cm} \\ 
    \end{tabular}
    \caption{\textbf{Detected Materials and Density as well as Percentage Error [\%] from the actual Physical Values of the Objects.} The measurements for Phone, Chair 2, Bicycle 1, Bench 2, and Car deviated significantly from the actual values. These deviations stemmed from incorrect detection of the material.}
    \label{tab:DetectabilityofMaterials}
\end{table}

However, the FMD dataset only achieves an accuracy of 52 $\%$ across all classes. One of the reasons is the insufficient data set, which consists of only ten categories in the neural network and 23 in the MINC data set. Increasing the data density would significantly increase the hit rate. With more extensive training, our approach can also be transferred to other environments. This requires the RGB image for analysis and the database with the necessary 3D networks and recognition models for evaluation.

The actual size of objects within a scene has not been considered in previous work. This would be useful for volume calculation and different scaling of objects. Even within the categories, no distinctions are made between different types of objects. Each object is only assigned a 3D mesh, which is considered the average for that class. This means that object shapes are not taken into account. Furthermore, for successful mapping and analysis of physical properties, the use of error-free, detailed, and complete 3D models is essential. However, depending on the orientation and movement of the objects, the calculated volume may be inaccurate. Serial images could help alleviate this error.

\section{Limitations}
While our implementation shows promising initial results, it solely serves to illustrate the feasibility of our proposed concept. Although our implementation used neural networks that were trained with real-world images, we relied on synthetic datasets to assess the performance of the implementation. Consequently, the generalizability and applicability of our findings may be limited.

Additionally, our evaluation of the detected physical properties examined the average density of each object. Therefore, our work might not guarantee that our approach is readily applicable to other properties, such as plasticity or thermal conductivity, as well as to more complex object compositions.

\section{Conclusion and Future Work}

In this paper, we presented a concept for the object-based recognition and assignment of physical properties as density or material based on a 2D image. 
Our work is motivated by the challenges of distinguishing objects and their properties from each other. The distinction of mass or density enables new interaction possibilities in which the causal relationships of an environment can be linked to the properties of a given object.  
Our method recognizes specific patterns from 2D images by neural networks in which we estimate the volume by the number of object-recognized triangles. The density is ultimately calculated from the object-specific assignment of a material recognition model and the associated volume. 

Despite the promising results of our approach, further work and improvements are needed. 
An essential aspect relates to the data sets that are used. The accuracy of object-based density recognition goes hand in hand with the quality of the trained AI model. Our current field of application is limited to synthetic test data. Future iterations and evaluations using real-world data sets could help deliver further insights into AI-based density recognition. To achieve serviceable recognition and acceptable results, the appropriate data set needs to be selected for the application area. Therefore, it is necessary to extend the data sets and the training model. The COCO dataset covers numerous categories but neglects existing subcategories. By selecting such data sets, the transferability of the model could be increased. 

Our evaluation shows only a limited number of materials assigned to the objects. In reality, the number and composition of materials may be different and more diverse (see Fig.~\ref{fig:MaterialComposition}). In particular, the interior composition of a recognized object may differ from the recognized surface texture. Consequently, drawing on and combining a wide variety of databases could lead to more precise and serviceable results. These assignments can be linked using an ontological approach. Utilizing ontologies, information and their relationship to each other can be stored in a machine-readable form or made comprehensible. Additional graph databases can visualize the data nodes and their relationship and make them interpretable. In principle, different ontologies can be merged within one ontology. In this context, possible databases on physical properties such as material, geometry and objects could be linked ontologically and applied to the principle of AI-based density recognition.  

Supporting these classifications with suitable image segmentation, such as with self-organized maps \cite{MM22}, could further increase the number of distinguishable materials. The partial change of the segmentable areas could be cut out or reduced to densely recognizable areas, which would also reduce quality restrictions and latencies.

%Sinn/Ausblick: Datenbanken verbinden, Objekte bestehen aus mehreren Materialien. Außerdem Inhalt, nicht nur Oberfläche

Estimating the object size within a scene proves to be a difficult task. This process assumes the same size for all everyday objects in a scene. In the future, it will be necessary to measure the object size and distance of acquired 3D models for meter-level distinctions. The model transfer to a spatial data set would be suitable for this purpose. 

The use of image series or video material can also be helpful to support a spatial data set~\cite{EKOA.24}. In this context, the distance and perspective of objects within a scene can be used to determine the speed and possible acceleration of an object. Javadi et al. \cite{JDP19} describe a video-based vehicle speed system for measuring speed based on a measured route. By determining the speed and acceleration of an object, statements can be made about the forces released in a collision. Video analysis can also be useful for other areas of physical property recognition. The depiction of an object in several individual images with different perspectives allows properties derived previously to be checked and re-evaluated. This includes, for example, the volume or size of the object.

\section{Acknowledgements}
We thank Philip Raschdorf for his support during concept development and data collection. We also thank Thomas Odaker and Elisabeth Mayer, who supported this work with helpful discussions and feedback.

%%%%%%%%%%%%%%%%%%%%%%%%%%%%%%%%%%%%%%%%%%%%%%%%%%%%%%%%%%%%%%%%%%%%%%%%%%%%%%%%%%%%%%%%%%%%%%%%%%

%\bibliographystyle{splncs04}
%\bibliography{main}

\end{document}